\documentclass[letterpaper]{article} 
\usepackage[draft]{aaai25}
\usepackage{times}  
\usepackage{helvet}  
\usepackage{courier}  
\usepackage[hyphens]{url}  
\usepackage{graphicx} 
\usepackage{natbib}  
\usepackage{caption} 
\usepackage{algorithm}
\usepackage{algorithmic}

\usepackage{newfloat}
\usepackage{listings}
\DeclareCaptionStyle{ruled}{labelfont=normalfont,labelsep=colon,strut=off} 
\floatstyle{ruled}
\newfloat{listing}{tb}{lst}{}
\floatname{listing}{Listing}

\usepackage[utf8]{inputenc} 
\usepackage{booktabs}       
\usepackage{amsfonts}       
\usepackage{nicefrac}       
\usepackage{microtype}      

\usepackage{subfig}

\usepackage{svg}
\usepackage{amsmath}
\usepackage{amssymb}
\usepackage{amsthm}

\theoremstyle{definition}
\newtheorem{definition}{Definition}
\newtheorem{theorem}{Theorem}

\usepackage{fixme}
\fxsetup{
  status=draft,
  inline,
  theme=color,
  nomargin,
  silent,
}
\fxsetup{marginface=\linespread{1}\footnotesize}
\definecolor{fxtarget}{rgb}{0.4,0.4,0.6}
\FXRegisterAuthor{a}{ae}{\color{teal} {\textbf{Arthur}}}
\FXRegisterAuthor{j}{je}{\color{orange} {\textbf{Jo\~ao}}}
\FXRegisterAuthor{l}{le}{\color{blue} {\textbf{Lucas}}}

\newcommand{\tento}[1]{\text{\footnotesize $\times10^{#1}$}}

\newcommand{\x}{\mathbf{x}}
\newcommand{\y}{\mathbf{y}}

\renewcommand{\H}{\mathcal{H}}
\newcommand{\X}{\mathcal{X}}

\newcommand{\R}{\mathbb{R}}

\title{Neural Conjugate Flows: \\Physics-informed architectures with flow structure}

\author {
    Arthur Bizzi\textsuperscript{\rm 1},
    Lucas Nissenbaum\textsuperscript{\rm 1},
    João M. Pereira\textsuperscript{\rm 1}
}
\affiliations {
    \textsuperscript{\rm 1}Instituto de Matemática Pura e Aplicada (IMPA), Rio de Janeiro, Brazil\\
    arthur.bizzi@impa.br, nissenbaum@impa.br, jpereira@impa.br
}

\begin{document}

\maketitle

\begin{abstract}
We introduce Neural Conjugate Flows (NCF), a class of neural-network architectures equipped with exact flow structure. By leveraging topological conjugation, we prove that these networks are not only naturally isomorphic to a continuous group, but are also universal approximators for flows of ordinary differential equation (ODEs). Furthermore, topological properties of these flows can be enforced by the architecture in an interpretable manner. We demonstrate in numerical experiments how this topological group structure leads to concrete computational gains over other physics informed neural networks in estimating and extrapolating latent dynamics of ODEs, while training up to five times faster than other flow-based architectures.
\end{abstract}

\section{Introduction}

The introduction of Physics-Informed Neural Networks (PINNs) \cite{raissi2019physics} has sparked interest in using neural networks to solve and discover differential equations. By encoding modeling into ``Physics-informed'' losses, networks $\mathcal{N}_\theta$ are trained to approximate \textit{solutions} to differential equations:
$$
\frac{d}{dt}\mathcal{N}_\theta(\x^0,t) = F\left(\mathcal{N}_\theta(\x^0,t)\right)
$$
$$
\iff \mathcal{L}(\theta) = \left|\left|\frac{d}{dt}\mathcal{N}_\theta(\x^0,t) - F\left(\mathcal{N}_\theta(\x^0,t)\right)\right|\right| = 0,
$$
where $F: \R^n \mapsto \R^n$ is a vector field, $\x^0 \in \R^n$ are initial conditions and $\mathcal{N}_\theta: \R^{n+1}\mapsto\R^n$ is a $\theta-$parameterized NN. $\mathcal{L}(\theta)$ denotes an equation loss relative to $\theta$.

PINNs have since been applied to forward and inverse problems across a wide range of domains \cite{de2022weak,patel2022thermodynamically,mao2020physics,rao2021physics,Ali1}.

Still, due to their black-box nature, PINNs struggle significantly with enforcing fundamental structural properties of the solutions. 
Thus, it is often necessary to encode constraints into PINNs through the use of additional Physics-based \textit{losses}. These might include terms that reward proper conservation of Hamiltonians \cite{mattheakis2022hamiltonian}, Lagrangians \cite{cranmer2020lagrangian} and symmetries \cite{zhang2023enforcing}.
 
PINNs are specially notorious for their issues with causal dependence on initial conditions, which manifest in the form of incorrect initial conditions or non-physical convergence to trivial solutions (see Fig. 1 for an example). Attempts have also been made to minimize these effects with modified losses, emulating causality \cite{wang2022respecting} or penalizing large gradients \cite{yu2022gradient}. 

In contrast, few attempts have been made to tackle these issues with new physics-based \textit{architectures}. Indeed, the vast majority of PINNs use general-purpose feed-forward architectures, such as the simple Multi-Layer Percetron (MLP) \cite{cuomo2022scientific}.

A remarkable solution to these causality issues may be found in the framework of Neural Ordinary Differential Equations (Neural ODEs) \cite{chen2019neural}. This approach is instead based on modeling the \textit{derivative} term of an ODE with a neural network:
$$\frac{d}{dt}{\mathbf{x}} = \mathcal{N}_\theta(\mathbf{x})$$
This ODE may then be solved with numerical solvers, using the adjoint method to calculate loss gradients.

This blend of neural and traditional methods leads to these networks having the structure of a \textit{flow}, which we detail in Section~\ref{sec:flows_conjugation}. This leads to a solution that is automatically compliant to initial conditions and that implicitly satisfies causality and time reversibility, making Neural ODEs specially adequate for physics-informed contexts \cite{lai2021structural,o2022stochastic}. 

However, the applicability of Neural ODEs is limited by the large computational overhead introduced by the sequential numerical solvers they are built upon. In particular, calculating gradients with the adjoint method leads to significant slowdowns \cite{kidger2022neural}. Regularizing and optimizing this process is an open problem \cite{finlay2020trainneuralode}.

\begin{figure*}[t!]
    \begin{minipage}[t]{.45\textwidth}
        \centering
        \includegraphics[width=\textwidth]{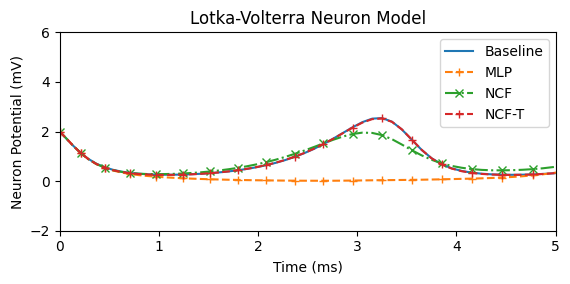}
        \caption{Spurious convergence of an MLP to equilibrium, denoting lack of uniqueness (see the Appendix for experimental details). The tighter structure of NCFs makes this less likely.}\label{fig:1}
    \end{minipage}
    \hfill
    \begin{minipage}[t]{.45\textwidth}
        \centering
        \includegraphics[width=\textwidth]{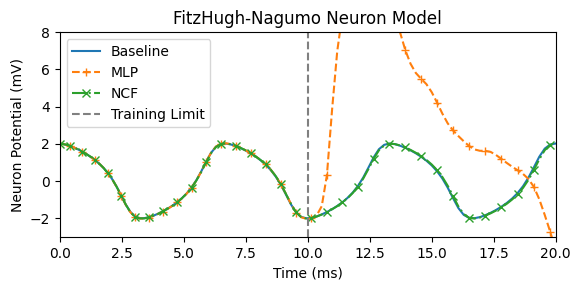}
        \caption{Extrapolation capacity of NCF versus an MLP (see Section 4 for details). The NCF is able to generalize beyond the training time of $10$ms, the MLP is not.}\label{fig:2}
    \end{minipage}  
    \label{fig:1-2}
\end{figure*}

A computationally lighter alternative to Neural ODEs was proposed in \cite{bilovs2021neural}. However, this approach comes at the cost of the associative property of flows (see Section 2), leading to an incomplete group structure. In particular, although this approach may lead to correct initial conditions, it cannot enforce unicity or causality. 

\subsection{Our contributions}

We propose a new flow-structured architecture, named \textit{Neural Conjugate Flows} (NCFs). These models reproduce the structure of solution groups for differential equations exactly,

by leveraging the dynamical-systems' framework of \textit{topological flow conjugation}. To the best of our knowledge, the only other instance of conjugation being applied to machine learning is in the context of approximating Poincaré sections of flows \cite{bramburger2021deep}, while \citet{neuralodewarping} use it indirectly to warp trajectories of Neural ODEs .

In essence, NCFs attempt to topologically deform nonlinear vector fields into those associated with integrable systems. This deformation takes place as a conjugation mapping parameterized by an invertible neural network $\mathcal{H}_{\theta}$. 
As a result, we have that:
\begin{itemize}
    \item NCFs are flow operators, and as such automatically respect initial conditions and time causality.
    \item NCFs with conjugate affine flows are universal approximators for flow operators of autonomous ODEs.
    \item Several topological properties of the solution of the ODE may be enforced through properly choosing the inner flow operator.
\end{itemize}

In this sense, NCFs may be seen as topology-informed alternatives to Neural ODEs. Though both architectures share similar objectives, NCFs may be trained significantly faster. This is due to the fact that the simpler flows used here may be calculated in closed-form in a parallel fashion, as opposed to the intrinsically sequential calculation of numerical flows.

  The paper is structured as follows.
  In Section~\ref{sec:flows_conjugation}, we revisit the notion of a flow operator, along with the notion of conjugation and affine flows.
  In Section~\ref{sec:NCFs} we introduce the general framework of Neural Conjugation and show that NCFs are universal approximators for flows of ODEs.
  Finally, in Section~\ref{sec:experiments} we present numerical experiments, 
  and in Section {5} we present some limitations of the architecture, along with final remarks. 

  In the supplemental materials, we have included an Appendix with additional experiments, proofs and implementation details.

\section{Flows and conjugation}\label{sec:flows_conjugation}

\subsection{Flows}\label{sec:flows}

Consider the prototypical system of autonomous ordinary differential equations in $\x\in \X \subset \mathbb{R}^n$:
\begin{equation} \label{ODE}
    \frac{d}{dt}{\x } = F(\x),\;\; \x(0) = \x^0 
\end{equation}
Now let $\Phi:[0,T]\times \X \mapsto \X$ be the \textit{flow operator} (or simply \textit{flow}\footnote{Note that invertible `flow' models, like Normalizing Flows \cite{papamakarios2021normalizing}, are inspired by mathematical flows, but in general do not have true flow structure.}) associated with the vector field $F$, defined as the map that propagates a initial state $\x$ through the ODE's dynamics for a given time interval $t$:
\begin{equation} 
    \Phi(t; \x^0) := \Phi^{t}\x^0 = \x(t)
\end{equation} 

It is a fundamental result in the theory of ODEs \cite{viana2021differential} that this operator should be not only continuous, but also behave as a \textit{group}, as encoded in the following group properties:

\textbf{I - {Identity Element:}}
\begin{equation} 
    \Phi^0 (\x) = \x,\quad \forall \x \in \X
\end{equation} 

\textbf{II - {Associativity:}}
\begin{equation} 
     \Phi^{t}\circ \Phi^{\tau}(\x) = \Phi^{t + \tau}(\x), \quad \forall \x\in \X,\, t,\tau \in \mathbb{R}
\end{equation} 
Together, \textbf{I} and \textbf{II} imply that flows are \textit{invertible}:

\textbf{``III'' - {Invertibility:}}
\begin{equation} 
    \Phi^{-t} \circ  \Phi^{t}(\x) = \x,\quad \forall \x\in \X,\, t \in \mathbb{R}
\end{equation} 

In many ways, these properties encode the notion of causality and well-posedness in physical systems:
\begin{itemize}
    \item Systems with property \textbf{I} respect initial conditions;
    \item Systems with property \textbf{II} respect time causality and uniqueness of trajectories;
    \item Systems with property \textbf{III} enjoy time reversibility.
\end{itemize}


For MLP-PINNs, none of these properties are guaranteed; instead, they must be soft-enforced by means of different and often conflicting Physics-informed losses.

\subsection{Topological Conjugation}\label{sec:conjugation}
Take two open subsets of $\mathbb{R}^n$, $\X_\Phi$ and $\X_\Psi$. Two flows $\Phi$ and $\Psi$ are said to be (locally) \textit{topologically conjugated} if there is a homeomorphism $H:\X_\Phi \to \X_\Psi$ such that:
\begin{equation} 
   \Phi^{t} (\x) =  H^{-1} \circ \Psi^{t} \circ H(\x), \quad \forall \x \in \X_\Phi,\,t\in \mathbb{R}
\end{equation}
where $H^{-1}$ is the inverse of $H$. Intuitively, conjugated systems are related by a change of variables. Moreover, given a flow $\Psi^t$ and a homeomorphism $H$, the conjugate operator $\Phi^{t} = H^{-1} \circ \Psi^t \circ H$ is also a flow: 

\textbf{I - {Identity Element:}}
\begin{equation} 
\begin{aligned}
        \Phi^0 (\x) &= H^{-1} \circ \Psi^0 \circ H (\x) \\
        &= H^{-1} \circ H (\x) = \x.
\end{aligned}
\end{equation} 

\textbf{II - {Associativity:}}
\begin{equation} 
\begin{aligned}
   \Phi^{t} \circ \Phi^{\tau} (\x) &=  H^{-1} \circ \Psi^{t} \circ H \circ H^{-1} \circ \Psi^{\tau} \circ  H (\x) \\ 
   &=  H^{-1} \circ \Psi^{t} \circ \Psi^{\tau} \circ  H (\x) \\
   &=  H^{-1} \circ \Psi^{t+\tau} \circ  H (\x) = \Phi^{t+\tau} (\x). \\
\end{aligned}
\end{equation} 
We may then use conjugation to construct new flows from known ones, as per the commutative diagram:


\begin{figure}[h!]
  \centering
  \includegraphics[width=.3\linewidth]{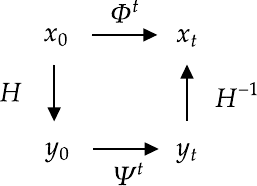}
  \label{fig:commutative}
\end{figure}

Fundamental theorems in the theory of dynamical systems leverage conjugation \cite{viana2021differential}:


\textbf{Flow-Box} or \textbf{Tubular Flow Theorem:} Near a regular point $p$ (one far from equilibria or cycles), a dynamical system is locally conjugated to a translation.

\textbf{Hartman-Grobman Theorem:} Near a hyperbolic equilibrium point $p$, a dynamical system is locally conjugated to its linearization around $p$ (See Fig. A in the Appendix).

Underlying these theorems is the fact that the vector fields associated with two conjugated flows are \textit{topologically equivalent}: orbits in their phase spaces may be continuously deformed into one another. In particular, cycles and equilibria are preserved:
\begin{figure}[h!]
  \centering
  \includegraphics[width=.3\linewidth]{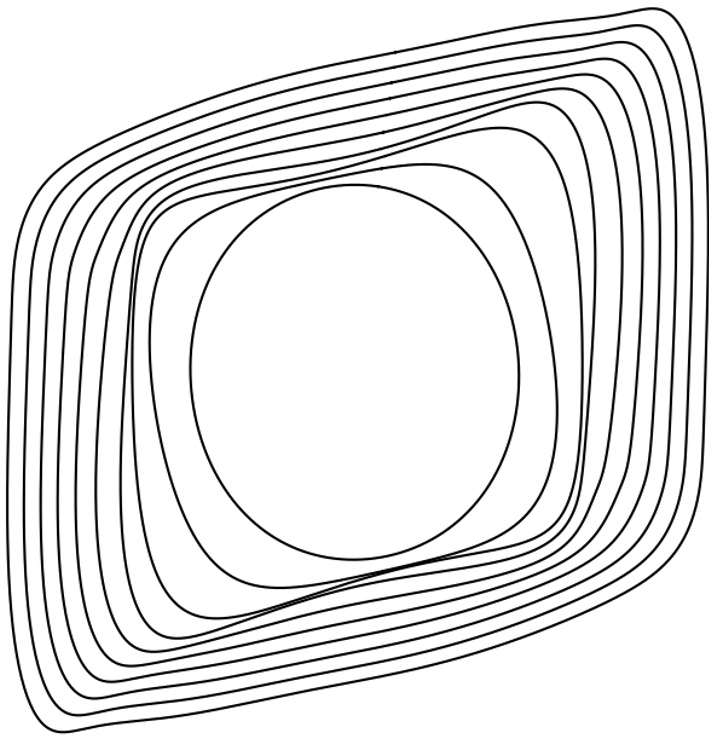}
  \caption{Deforming a harmonic oscillator's orbit to match the limit cycle of a Van-der-Pol system.}
  \label{fig:Deformation}
\end{figure}

 We may leverage this by controlling the structure of the conjugate $\Psi$. In this work, we will use the fact that matrix exponentials are \textit{Lie groups}, and thus have remarkable topologies \cite{Humphreys1972}. We can then, for example, obtain the topology of a sphere by conjugating with the the orthogonal group SO($n$).

\subsection{Affine Flows and Universality }\label{sec:topological_equivalence}

We now restrict ourselves to affine systems. Let $\mathbf{A} \in \mathbb{R}^{n\times n}$ and $\mathbf{b} \in \mathbb{R}^{n}$. We have:
\begin{equation} \label{AffineODE}
    \frac{d}{dt} \x = \mathbf{A}\x + \mathbf{b},\;\; \x(0) = \x^0,
\end{equation} 
Their associated flows are well known, and may be given in terms of matrix exponentials\footnote{In the Appendix, we give an alternate numerical formula that does not involve calculating integrals.}:
\begin{equation}
    \x(t) = e^{\mathbf{A} t} \x + \int_{0}^t e^{\mathbf{A} \tau} \mathbf{b}\,d\tau
\end{equation}
While this class of flows may at first seem too simple to be representative, we can leverage augmentation (or padding) to prove that any Lipschitz-continuous system of ODEs can be conjugated to an affine system. 

To augment a system is to equip it with additional ``dummy'' dimensions, allowing wider, more representative models to be used \cite{dupont2019augmentednode}. The increase in representation power enjoyed by systems in higher dimensions will be used both in our proofs and our implementation.

We will prove that any sufficiently well-behaved nonlinear system may be embedded in a larger dimensional manifold where it is essentially linear:
\begin{theorem}\label{thm:universality}
Let $F:\mathbb{R}^n\mapsto \mathbb{R}^n$ be a Lipschitz-continuous vector field. Then for any positive integer $m$ there exists an augmentation $\mathbf{a} \in \mathbb{R}^m$, a component $G: \mathbb{R}^n\times\mathbb{R}^m \mapsto \mathbb{R}^m$, a matrix $\hat{\mathbf{A}} \in \mathbb{R}^{(n+m)\times(n+m)}$ and a vector $\hat{\mathbf{b}}\in \mathbb{R}^{n+m}$ such that the following augmented system
\begin{equation}\label{eq:augmented_system}
    \frac{d}{dt}{\hat{\x}} = \frac{d}{dt}{\begin{bmatrix} \x \\ \mathbf{a}\end{bmatrix}} = \begin{bmatrix} F(\x) \\ G(\x, \mathbf{a}) \end{bmatrix} = \hat F(\hat{\x})
\end{equation}
 is conjugated to the affine system:
\begin{equation} 
    \frac{d}{dt}{\mathbf{y}} = \hat{\mathbf{A}} \mathbf{y} + \hat{\mathbf{b}}.
\end{equation}
\end{theorem}

In other words, we may extend nonlinear systems to a higher dimension in a way that unravels their orbits, so they can then be deformed to match the orbits of an affine system. This result can be thought as an extension of the Tubular-Flow Theorem: In essence, we may always choose an embedding that ``destroys'' the system's topology, making every point regular.

The proof, which we give in full in Appendix A, is constructive, using the flow operator of the ODE to construct the homeomorphism $H$. 

This means that we may solve any sufficiently mild ODE as a part of a larger solution of the form:
\begin{equation}
    {\begin{bmatrix} \mathbf{x}(t) \\ \mathbf{a}(t)\end{bmatrix}} = H^{-1} \left( e^{\hat A t} H\left ( {\begin{bmatrix} \mathbf{x}^0 \\ \mathbf{a}^0\end{bmatrix}} \right) + \int_{0}^t e^{\hat A \tau} \hat b\,d\tau\right).
\end{equation}

\begin{figure*}[t]
  \centering
  \includegraphics[width=.55\linewidth]{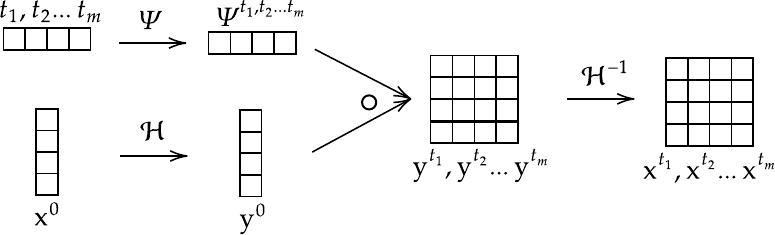}
  \caption{The NCF pipeline: Change variables to the conjugate manifold, iterate, then change back.}
  \label{fig:NCF_pipeline}
\end{figure*}

\section{Neural Conjugate Flows}\label{sec:NCFs}

\subsection{Neural conjugation}


\begin{definition}[Neural Conjugate Flow] A \textit{Neural Conjugate Flow\footnote{Note that although time-one maps for NCFs could be used to implement normalizing flows, they are are two distinct concepts.}} $\Phi^t$ is defined as:
\begin{equation}
\Phi^t = \mathcal{H}^{-1} \circ \Psi^t \circ \mathcal{H},
\end{equation} 
where $\Psi$ is a flow operator, $\mathcal{H}$ is an invertible neural network and $\mathcal{H}^{-1}$ is its inverse.
\end{definition}

In principle, any invertible architecture could be used to approximate $\mathcal{H}$. For this paper, we have chosen the 'Coupling Layer' architecture, an Universal Approximator for homeomorphisms (see section 3.2).

Likewise, the flow $\Psi$ could be either analytical or numerical in nature. For this paper, we will use the analytical affine flow, which is an universal approximator under conjugation, as discussed (see also section 3.3).

Inference in an NCF then takes place in three steps:
\begin{enumerate}
    \item We feed the initial condition $\x^0$ to the network $\mathcal{H}$: 
    $$\y^0 = \mathcal{H}(\x^0).$$
    \item We apply the flow $\Psi$ evaluated at times $t_1, t_2, \dots, t_m$:
    $$\y^{t_i}=\Psi^{t_i} (\y^0),\quad i=1,\dots,m.$$
    \item We feed the result to the inverse network  $\mathcal{H}^{-1}$:
    $$\x^{t_i}=\mathcal{H}^{-1}(\y^{t_i}),\quad i=1,\dots,m.$$
\end{enumerate}
A schematic of this procedure is presented in Figure ~\ref{fig:NCF_pipeline}. A detailed view of our implementation may be found in Appendix B.

The combined universality of affine flows and Coupling Layers allows us to prove our main theorem: Affine Neural Conjugate Flows are universal approximators for flows of ODEs.

\begin{theorem}[Universal Approximation of Affine NCFs]\label{thm:universality2}
Let $\Phi$ be a flow associated with a differentiable vector field. There are an augmentation $\hat\Phi$ and a Neural Conjugate Flow $\Tilde \Phi = \mathcal{H}^{-1} \circ \Psi^t \circ \mathcal{H}$, with $\mathcal{H}$ a 
Coupling Layer ensemble and $\Psi$ affine, such that $\Tilde\Phi$ approximates $\hat\Phi$.
\end{theorem}

 The proof is straightforward (see Appendix A). This result indicates that, given a sufficiently representative Coupling Layer, along with a properly crafted augmentation, NCFs can solve any differential equation where global existence is assured.

\subsection{The network $\mathcal{H}$}

In principle, any invertible architecture could be used to represent $\mathcal{H}$. This includes approximately invertible ones, like AutoEncoders \cite{kingma2022autoencoding} and U-nets \cite{ronneberger2015unet}. In practice, however, we have not been able to make neural conjugates based on these models converge. We conjecture that having the group properties of flows \textit{during} training (as opposed to after it) enhances convergence. 

\begin{figure}[t]
  \centering
  \includegraphics[width=.85\linewidth]{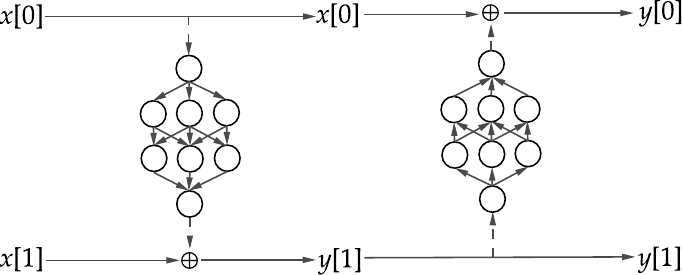}
  \caption{A standard coupling layer ensemble in $\mathbb{R}^2$. Inverse evaluation may be done by reversing the arrows and replacing $+$ for $-$. Notice that each MLP only `sees' half the input, leading to reduced representation power.}
  \label{fig:CL}
\end{figure}

Our architecture of choice for $\mathcal{H}$ is a composition of Coupling Layers \cite{dinh2016normalizingflows} . 
These networks are universal approximators for diffeomorphisms \cite{teshima2020coupling}, whose inverse is easily computable. 
Coupling Layers achieve perfect invertibility by applying neural networks to a split input, keeping half the dimensions constant at each layer (see Fig. \ref{fig:CL}). This comes at the cost of reduced representation power, as states are weakly coupled \cite{draxler2024universality}.

\begin{figure}[h!]
  \centering
  \includegraphics[width=.8\linewidth]{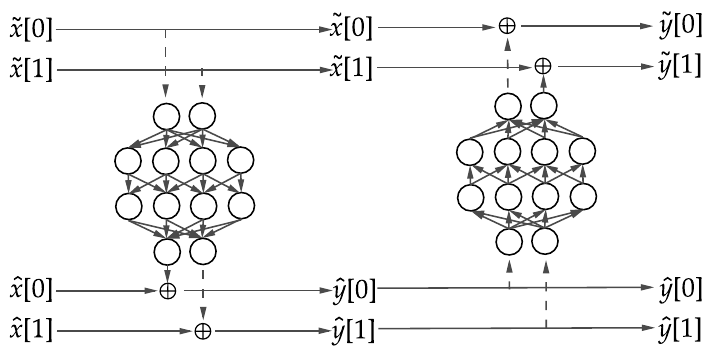}
  \caption{Our coupling layer ensemble. By twinning the input, we ensure each MLP sees both positions of the $\x$ vector, $\x[0]$ and $\x[1]$. }
  \label{fig:CL_ours}
\end{figure}

\begin{figure*}[t]
  \centering
  \includegraphics[width=.65\linewidth]{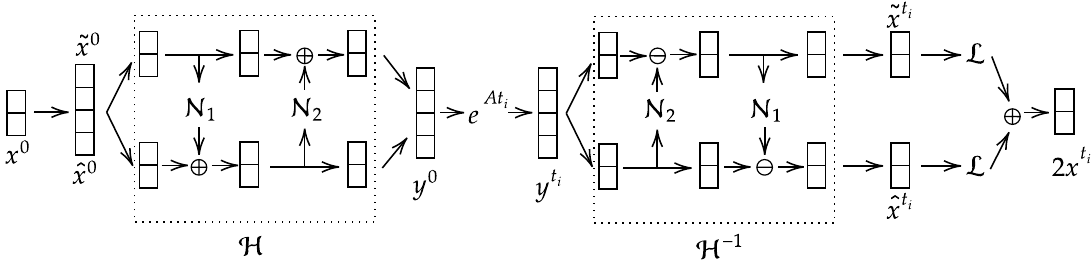}
  \caption{Our augmented Affine Neural Conjugate, in full. The following operations are applied in order, from left to right: input duplication, the homeomorphism $\H$ (comprised of two coupling layers), the affine flow $\Psi$ and the inverse homeomorphism $\H^{-1}$. The outputs $\Tilde{\x}^t,\hat{\x}^t$ are each trained according to the Physics-informed loss $\mathcal{L}$ then finally averaged.}
  \label{fig:Everything}
\end{figure*}

To recover expressiveness, we again turn to augmentation. We concatenate a copy of $\x^0$ to the input, widening our model by a factor of $2$ and allowing the Coupling Layers at each step to capture the entirety of the $n$-dimensional state $\x$ (see Fig. \ref{fig:CL_ours}). We then solve a duplicated ODE\footnote{See Fig (\ref{fig:Everything}) in a visual representation.}:
\begin{equation}
\frac{d}{dt} \begin{bmatrix}
        \Tilde{\x}\\\hat{\x}
    \end{bmatrix} = \begin{bmatrix}
        F(\Tilde{\x})\\F(\hat{\x})
    \end{bmatrix}
\end{equation}
and take the output to be the average  $\x = (\Tilde{\x} + \hat{\x})/2$.

In addition to using more expressive invertible layers, augmentation allows us to use affine flows in $2n$ dimensions, leading to substantially more expressive conjugates $\Psi$. Moreover, by projecting onto a higher-dimensional space we aim to trigger the conditions for universality (see Theorem 1).

 This comes at the cost of managing the additional `twin' dimensions, as conjugation is necessarily dimension-preserving. Note that both $\Tilde{\x}$ and $\hat{\x}$ must be trained as legitimate solutions to the problem; otherwise, if one of them is allowed to drift freely (as is done in e.g. the ANODEs framework \citep{dupont2019augmentednode}), the averaging projection would lead to trajectory crossing and the flow structure would be lost. 
 
 Empirically, this scheme can be interpreted as follows: although our flows are taking place in a $2n-$dimensional space, during training we attempt to restrict our dynamics to the codimension-$n$ `diagonal' manifold $(\x,\x)$, where flow structure is preserved.

\subsection{The flow $\Psi$}

Our architecture is best suited for systems for which some qualitative description is known. Theoretical or numerical results may provide insight into a system's topology, which can then be leveraged to choose a suitable conjugate. Explicit conjugates, although rare, should make for excellent flows $\Psi$. 

Motivated by Universality theorem 1, we have set our conjugate flows $\Psi$ to be affine flows. We then let the parameters $\mathbf{A}$ and $\mathbf{b}$ of $\Psi$ be trained along with the parameters of $\mathcal{H}$. Although the proof (see Appendix A) seems to imply that purely translational flows ($\mathbf{A}=0$) suffice for universality, allowing for a learnable matrix $\mathbf{A}$ seems to perform better in practice.

Affine flows may be calculated very quickly, at least for systems in low dimensions: The exponentials $\exp(\mathbf{A}t_i)$ may be calculated entirely in parallel using batching/broadcasting, in contrast to numerical solvers.

Additionally, we may enforce topological properties of the flow by appropriately projecting $\mathbf{A}$ onto a suitable matrix algebra, defining the topology of the associated Lie group through the exponential map. For example, if we desire oscillatory behavior, we may enforce the topology of the orthogonal group by making $\mathbf{A}$ the skew-symmetric component of a trainable parameter $\mathbf{M}$: 
\begin{equation} 
       \mathbf{A} = (\mathbf{M} - \mathbf{M}^T)/2.
\end{equation}

We interpret our architecture as approximating a flow by that of an \textit{integrable} system. As a result, Neural Conjugates of affine flows should perform poorly when emulating strongly nonlinear features, such as convergence towards limit cycles and strange attractors. We show that this is indeed the case in Appendix C.

Moreover, this indicates that careful initialization of the matrix $\mathbf{A}$ and the vector $\mathbf{b}$ are necessary to achieve convergence. We have observed that a poor initialization leads to a tug-of-war between $\Psi$ and $\mathcal{H}$. To avoid this, we initialize $\mathcal{H}$ near the identity and $\Psi$ as a suitable affine approximation to the system dynamics around $\x^0$ (see Appendix B).
\begin{equation}
    \mathbf{A}_0\x^0 + \mathbf{b}_0 \approx F(\x^0).
\end{equation}

\newpage 
\section{Numerical Experiments}\label{sec:experiments}

We will validate our models by solving forward and inverse problems in the realm of mathematical neuroscience. We aim to evaluate how well they can learn the firing/spiking dynamics of different biological neuron models, with and without data.

We evaluate three metrics. The first metric is accuracy, measured in terms of the mean-squared error between the estimate obtained by each method and the ground truth baseline, obtained using a high-accuracy numerical solver. Our second metric is total training time. Our third metric is the models' extrapolation capacity, through which we aim to establish if the model has learned to generalize the system's dynamics. To measure this, we compute the model beyond the training interval $[0,T]$, computing its accuracy relative to the ground truth over the interval $[0,2T]$.

The baseline MLP-PINNs were built according to current best practices \cite{wang2023expert}, including a Gaussian Fourier feature layer \cite{tancik2020fourier} with $\sigma=2$. We use the initial condition trick in \cite{bilovs2021neural}, Xavier initialization \cite{kumar2017weightxavier} and tanh activations. 

For the Neural ODEs, we used the Pytorch implementation given by the TorchDyn library \cite{politorchdyn}. This library offers fully optimized implementations for the networks and the associated numerical solvers. A second-order, semi-implicit (midpoint) solver with fixed time-step was used to prevent the slowdowns of up to one order of magnitude imposed by higher-order methods. The vector fields were represented by MLPs with tanh activations and Xavier initialization. 

Finally, we evaluate two implementations of Neural Conjugate Flows: vanilla, denoted NCF, and topologically enforced, denoted NCF-T. The homeomorphisms $\H$ are composed of two MLP-based Coupling Layers.

Each model was trained for 2000 epochs, full-batch, and optimized with ADAM \cite{kingma2014adam}. Each experiment was run five times, after which the average and standard deviation of each metric were evaluated. They were executed on the same machine, equipped with an AMD Ryzen 9 5900HX processor, an RTX 3060 GPU and 16GB of RAM. Further specifications may be found in Appendix C.

\newpage 
\subsection{Forward Problem: FitzHugh-Nagumo}

The FitzHugh-Nagumo (FH) system \cite{FITZHUGH1961445} is a minimal description of the activation dynamics of a spiking neuron. It is a prime example of a relaxation oscillator and exhibits rich dynamics, including limit cycles, excitation blocks and anodal breaks. Additionally, it becomes stiff for small $\epsilon$. The equations read:
\begin{align}\label{eq:fitzhugh}
\frac{dV_m}{dt} &= V_m - \frac{V_m^3}{3} - r + I \\
\frac{dr}{dt} &= \epsilon (V_m + a - br)
\end{align}
where \( V_m \) is the neuron membrane potential, \( r \) is a recovery variable associated with sion channels, \( I \) is an external current, and \( \epsilon \), \( a \), and \( b \) are parameters.

We compare three models: MLP, NCF and NCF-T.As in this case we have perfect knowledge of the differential equation, making a comparison to Neural ODEs would be unfair: one might as well use a traditional numerical solver in their place.

For this experiment, we compare how well MLP-PINNs and NCFs can solve the FH system with \textit{no data}, using only the associated Physics-informed loss. 
For training, we sub-divided the time-interval uniformly in 100 samples $t_i$. 
 A single Physics-informed loss was used, implemented as an $L_2$ penalty for the ODE residue at $t_i$:
 \begin{equation} 
   \mathcal{L}_{\text{PINN}}(\theta) =  \frac{1}{100}\sum_{i=1}^{100}\left|\left| \frac{d}{dt}\mathcal{N}_\theta(\x^0,t_i) - F(\mathcal{N}_\theta(\x^0,t_i)) \right|\right|_2^2,
\end{equation}
where $\mathcal{N}$ stands for the model and $F$ stands for the right-hand side of (18) and (19). For evaluation, we measured the the accuracy and generalization losses as:
 \begin{equation} 
   \mathcal{L}_{\text{acc,extrap}}(\theta) =  \frac{1}{100}\sum_{i=1}^{100}\left|\left| \mathcal{N}_\theta(\x^0,t_i) - \x^{t_i} \right|\right|_2^2,
\end{equation}
where for measuring accuracy we take $0 \leq t_i \leq 10$, while for generalization we take $0 \leq t_i \leq 20$. The results are listed in Table 1. Experimental details may be found in Appendix C.

\subsection{Inverse Problem: Hodgkin-Huxley}

The Hodgkin Huxley model \cite{Hodgkin1952} is the canonical neuron model used in mathematical biology, based on the dynamics of the giant squid axon. The model is a fourth-order, nonlinear differential equation that displays remarkably intricate phenomena, including chaos \cite{guckenheimer2002chaos}. The equation for the neuronal voltage is given by: 
\begin{equation}
\begin{aligned}
 C_{m}{\frac {{\mathrm {d} }V_{m}}{{\mathrm {d} }t}}=&-I+{\bar {g}}_{\text{K}}n^{4}(V_{m}-V_{K})
 \\&+{\bar {g}}_{\text{Na}}m^{3}h(V_{m}-V_{Na})+{\bar {g}}_{l}(V_{m}-V_{l})
\end{aligned}
\end{equation}
where $Vm$ is the membrane potential, $C_m$ is the membrane capacitance, $I$ is the stimulus current, $V_K,V_{\text{Na}},V_l$ are the base voltages for sodium, potassium, and leak currents, $\bar{g}_\text{Na}, \bar{g}_\text{K}, \bar{g}_\text{L}$ are the respective maximal conductances and
   $E_\text{Na}, E_\text{K}, E_\text{L}$  are the respective reversal potentials.
   
   $m, h, n$ are gating variables for sodium activation, sodium inactivation, and potassium activation, which follow their own set of nonlinear equations, determined experimentally. 
For this experiment, we measure the networks' capacity to use both data \textit{and} physics to reproduce the dynamics of the neuron.  We attempt to emulate the experimental setup for finding surrogate models for $n,m,h$ from measurements \cite{johnston1994foundations} using a simulation as ground-truth. 

We compare four models: MLP, NCF, NCF-T and NODE. We sub-divide the time-interval $[0,14]$ into 100 equally spaced samples and train full-batch, now using two losses. 

First, a Physics-informed loss associated \textit{only} to the principled model for $V_m$ given in eq. (22):
 \begin{equation} 
   \mathcal{L}_{\text{PINN}}(\theta) =  \frac{1}{100}\sum_{i=1}^{100}\left|\left| \frac{d}{dt}\mathcal{N}_\theta(\x^0,t_i)[0] - F(\mathcal{N}_\theta(\x^0,t_i))[0] \right|\right|_2^2,
\end{equation}

Second, a data-driven loss based on noisy samples $\Tilde{\X}$ of the remaining three variables $n,m,h$:
 \begin{equation} 
   \mathcal{L}_{\text{data}}(\theta) =  \frac{1}{100}\sum_{i=1}^{100}\left|\left| \mathcal{N}_\theta(\x^0,t_i)[1:3] - \Tilde{\x}^{t_i}[1:3]\right|\right|_2^2,
\end{equation}

The evaluation metrics are the same as for Experiment 1. Our objective is to observe if the models can combine heterogeneous information into a solid model for the spiking dynamics. The result are in Table 2. Details may be found in Appendix C.

\begin{figure}[h!]
  \centering
  \includegraphics[width=.7\linewidth]{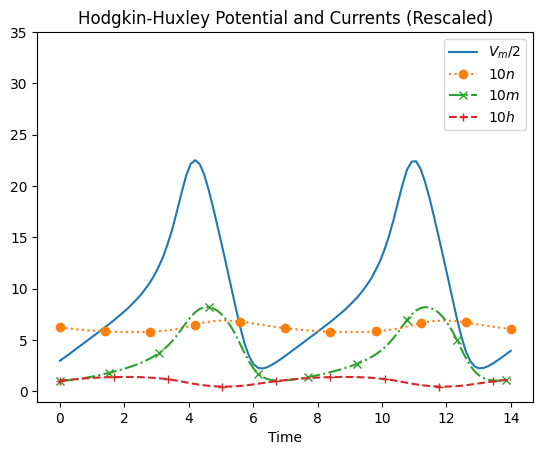}
  \caption{Reference values for the membrane potential and leak currents, rescaled for better visualization.}
\end{figure}


\begin{figure}[h!]
  \centering
  \includegraphics[width=.7\linewidth]{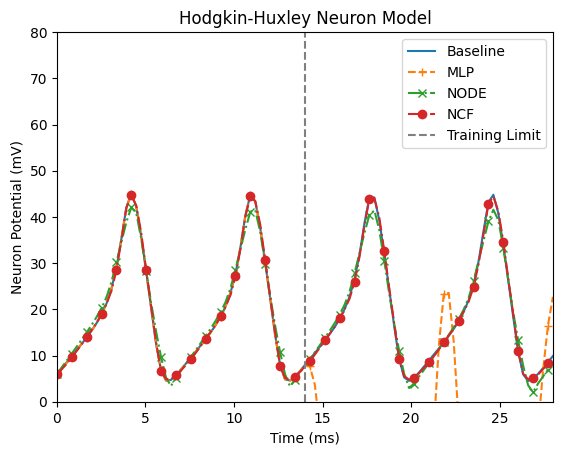}
  \caption{Reconstructed and extrapolated membrane potentials after training.}
\end{figure}

\begin{table*}[t]
\footnotesize
\addtolength{\tabcolsep}{-2pt}
  \caption{Experiment 1: FitzHugh-Nagumo model}
  \centering
  \begin{tabular}{ccccccc}
    \toprule
    Model & & Layers & Params   & $\mathcal{L}_{\text{acc}}$ & $\mathcal{L}_{\text{extrap}}$ &  Time(s)\\
    \midrule
    MLP & &$3\rightarrow32\rightarrow32\rightarrow32\rightarrow2$ & $2.3$K & $4.9\tento{-4} \pm 2.7\tento{-5}$ & $2.8\tento{1} \pm 5.4\tento{0}$ & $35.7\pm0.9$  \\
    NCF & & $2\times(2\rightarrow32\rightarrow32\rightarrow2)$ &$2.5$K & $6.7\tento{-2} \pm 1.0\tento{-2}$ & $2.5\tento{-1} \pm 3.0\tento{-2}$ & $74.7\pm0.5$  \\
    NCF-T & & $2\times(2\rightarrow32\rightarrow32\rightarrow2)$ &$2.5$K & $3.2\tento{-4} \pm 2.6\tento{-5}$ & $1.9\tento{-3} \pm 4.6\tento{-4}$ & $75.8\pm2.2$  \\
    \bottomrule
  \end{tabular}
\end{table*}

\begin{table*}[t]
\footnotesize
\addtolength{\tabcolsep}{-2pt}
  \caption{Experiment 2: Hodgkin-Huxley model}
  \centering
  \begin{tabular}{ccccccc}
    \toprule
    Model & & Layers &Params   & $\mathcal{L}_{\text{acc}}$ & $\mathcal{L}_{\text{extrap}}$ &  Time(s)\\
    \midrule
    MLP & & $5\rightarrow128\rightarrow128\rightarrow4$ &$17.0$K & $3.2\tento{-5} \pm 3.5\tento{-5}$ & $7.7\tento{0} \pm 1.7\tento{0}$ & $26.5\pm0.8$  \\
    NCF & & $2\times(4\rightarrow90\rightarrow90\rightarrow4)$ &$18.1$K & $1.8\tento{-2} \pm 4.4\tento{-3}$ & $3.2\tento{-2} \pm 3.2\tento{-1}$ & $44.4\pm1.9$  \\
    NCF-T & & $2\times(4\rightarrow90\rightarrow90\rightarrow4)$ &$18.1$K & $1.6\tento{-3} \pm 9.7\tento{-4}$ & $1.6\tento{-3} \pm 9.8\tento{-4}$ & $45.6\pm2.4$  \\
    NODE & & $4\rightarrow128\rightarrow128\rightarrow4$ &$17.7$K & $3.5\tento{-2} \pm 9.0\tento{-3}$ & $3.8\tento{-2} \pm 9.1\tento{-3}$ & $241.1\pm11.2$  \\
    \bottomrule
  \end{tabular}
\end{table*}

\subsection{Analysis and Additional Results}

We observe that in general, neural conjugates are slower than classical feed-forward architectures roughly by a factor of two, while being faster than Neural ODEs roughly by a factor of five. 

Our belief is that the speed deficit compared to MLPs can be mostly attributed to the fact that the input must flow through $\mathcal{H}$ twice (one for $\mathcal{H}$ and one for the inverse $\mathcal{H}^{-1}$). On the other hand, the speed advantage in comparison to Neural ODEs comes from the fact that the affine flows $\Psi$ may be calculated in an entirely parallel manner, as opposed to the sequential nature of the solvers built into Neural ODEs.

We observe that although MLP-PINNs can interpolate solutions quickly and accurately, they do not display any generalization power.  On the other hand, Neural ODEs and Neural Conjugates can be said to be truly learning the latent neuron dynamics, as illustrated by their extrapolation capabilities. We have also observed that NCFs both interpolate and extrapolate solutions better than Neural ODEs, at significantly lower computational cost.

The results show that NCFs perform particularly well when equipped with a well-chosen topology. There are situations for which we expect NCFs to perform worse than the other models, however. This is most evident when dealing with orbits which develop strongly nonlinear phenomena, such as chaotic behavior or complex limit cycles. We demonstrate this in Appendix C.

\section{Final Remarks}\label{sec:conclusion}

We have introduced Neural Conjugate Flows, a novel Physics-Informed architecture with the structure of flows, achieved via the composition of invertible neural networks and affine systems. Through the use of topological conjugation, these networks have the properties of continuous groups by construction, leading to automatic compliance with initial conditions and enhanced causality. It is also topology-informed.

Numerical experiments demonstrate that, despite the reduced representation power of invertible networks, our architecture consistently outperforms both Multi-layer Perceptrons and Neural Ordinary Differential equations at extrapolating the dynamics of latent dynamical systems. Moreover, they are up to five times faster than Neural ODEs, while being about twice as slow as conventional PINNs.

However, NCFs as implemented here are strongly limited by the representation capacity of the central flow $\Psi$. If it is chosen to be just affine, $\Psi$ deals poorly with systems with strongly nonlinear behavior. We believe that Neural Conjugation as a paradigm could be used in other contexts, by conjugating with more powerful and flexible flows.

It is in this sense that we believe that Neural Conjugates could flourish in the context of scientific machine learning, at the boundary of model-based numerical methods and ML.

\section*{Acknowledgments}

The authors would like to thank professors Sergey Tikhomirov and Marcelo Viana for their insights and suggestions.

\bibliography{aaai25}
\section*{Appendix}

\appendix 
\section{Universal Approximation}

To prove that Neural Conjugate Flows are Universal Approximators for solutions of ODEs, we need only find a class of flows $\Psi$ that are conjugated to any ODE and a class of invertible networks that can approximate any homeomorphism $H$. 

For $H$, the class of coupling layers suffices, as discussed. Thus, sufficiently large ensembles $\mathcal{H}$ are universal approximators for diffeomorphisms. 

As for $\Psi$, we may choose the class of affine flows, as we have established. We'll prove this next.

\subsection{Universality of Affine Conjugation}

We prove that every ordinary differential equation with global solutions may be embedded in a higher dimensional equation where it is conjugated to a translation. The proof is simple and gives an explicit construction for the embedding and for the homeomorphisms $H$ and $H^{-1}$.

In view of our application, we deliberately weaken the statement of the theorem, claiming there exists \textit{some} embedding that allows conjugation to \textit{some} affine system. We choose this wording because the actual implementation leaves the network free to choose both an embedding and a system that may better suit the target dynamics. What is important is that there is always \textit{at least one} such choice, which we provide here.

%
%
%

 \begin{proof}[Proof of Theorem 1]

We take the minimal augmentation in $m=1$ additional dimensions, with $a \in \mathbb{R}$ and  $G(a, \x):=1$. this leads to the augmented system: 
\begin{equation} 
     \frac{d}{dt}{\hat \x} =  \frac{d}{dt}{\begin{bmatrix} \x \\ a\end{bmatrix}} = \begin{bmatrix} F(\x) \\ 1\end{bmatrix} = \hat F(\hat \x)
\end{equation}
Obviously, if $F$ is Lipschitz continuous, so is $\hat F$. Furthermore, the variables $a$ and $\x$ are independent, which implies that the flow $\hat \Phi: \mathbb{R}\times\mathbb{R}^{n+1}\mapsto\mathbb{R}^{n+1}$, associated with $\hat F$, has a simple expression in terms of the flow $\Phi $ associated with $F$:
\begin{equation} 
    \hat \Phi^t\left(\begin{bmatrix} \x \\ a \end{bmatrix} \right) = \begin{bmatrix} \Phi^t x \\ a+t \end{bmatrix}
\end{equation}
We will prove that the flow $\hat \Phi$ is conjugated to the affine system with $\hat A=0$ (the all-zero $n+1 \times n+1$ matrix) and set $b = [0_n,1]^T$, (the vector with $n$ zeros and a one in its last position). The flow associated with this affine system is given by
\begin{equation} 
    \hat \Psi^t\left(\begin{bmatrix} \x \\ a \end{bmatrix} \right) = \left(\begin{bmatrix} \x \\ a + t \end{bmatrix} \right).
\end{equation}
We show the two flows are conjugated by providing an explicit formula for the conjugation $\mathcal{H}$ and its inverse:
\begin{equation}
    \mathcal{H}\left(\begin{bmatrix} \x \\ a \end{bmatrix} \right) = \begin{bmatrix} \Phi^{-a} \x\\ a \end{bmatrix} \quad
    \; \; \mathcal{H}^{-1}\left(\begin{bmatrix} \x \\ a \end{bmatrix} \right) = \begin{bmatrix} \Phi^{a}\x \\ a \end{bmatrix}.
\end{equation}
We note that $H$ and its inverse are indeed well-defined in $\mathbb{R}^n$. Indeed, Picard's Existence Theorem establishes that since $\hat F$ is globally Lipschitz, we have that the flow $\Phi^a$ and its time-reversal $\Phi^{-a}$ are  well-defined everywhere, which implies that $\mathcal H$ is well defined and a homeomorphism.
Finally, the proof follows by algebraic manipulation:
\begin{align} 
 \mathcal H^{-1}\circ \hat \Psi^t \circ \mathcal H \left(\begin{bmatrix} \x \\ a \end{bmatrix}\right) 
 &= \mathcal H^{-1}\circ \hat \Psi^t \left( \begin{bmatrix} \Phi^{-a}\x \\ a \end{bmatrix}\right) \\
 & = \mathcal H^{-1}\left(\begin{bmatrix} \Phi^{-a}\x \\ a + t \end{bmatrix}\right)\\
 & = \begin{bmatrix} \Phi ^{t+a}\Phi^{-a}\x \\ a + t \end{bmatrix} \\
 & = \begin{bmatrix} \Phi ^{t}\x \\ a + t \end{bmatrix} = \hat \Phi ^{t}\left(\begin{bmatrix} \x \\ a \end{bmatrix}\right).
\end{align} 

\end{proof}


\section{Implementation Details}


\subsection{Evaluating the flow of affine systems}\label{sec:eval_affine_flow}

A simple, well-known trick to evaluate an affine flow without calculating integrals may be given as follows. We again augment the system, adding a dummy state with no dynamics so that the resulting purely linear system behaves as an affine system:

\begin{equation}
    \frac{d}{dt}{\begin{bmatrix}
        \x \\ a
    \end{bmatrix}} = \begin{bmatrix}
        \mathbf{A} & \mathbf{b}\\
        0 & 0
    \end{bmatrix}
    \begin{bmatrix}
        \x \\ a
    \end{bmatrix}
\end{equation}
If we now set $a^0=1$, we get that the dynamics of $\x$ under this flow is now exactly that of an affine flow:

$$
 \frac{d}{dt}{\x} = \mathbf{A}\x + \mathbf{b}
$$

\subsection{Initialization}

We first define a naive initialization scheme for NCFs, then a more realistic one.

Affine NCFs fundamentally try to approximate the target dynamics by that of a higher-dimensional affine system. Given an initial position $\x^0$, a natural first attempt would be to try and fit a linear system passing by $\x^0$ such that its dynamics are as close as possible to those of the target system. We do this in three steps:

  \textbf{1. Set $\mathcal{H}$ to be close to the identity:} This can be done by multiplying the output of the neural networks used in the coupling layer ensemble by a small value.

  \textbf{2. Map equilibria to the origin:}  Locate an equilibrium close to $\x^0$, say at $\Bar{\x}$ and make a change of variables $\x \mapsto \x-\Bar{\x}$. If there are none, we just assume it to be located at $\x=0$.

  \textbf{3. Define $\mathbf{A}$:}  Compute the matrix $\mathbf{A}$ such that $\mathbf{A}\x$ best approximates the dynamics at $\x^0$.

We claim $\mathbf{A}$ is the best possible linear approximation at $\x^0$ if $\mathbf{A}\x^0$ exactly matches the first derivative and achieves the minimum possible error for the second. This may be expressed as:
  \begin{align}
      \min_A||A-\mathbf{J}_0|| \text{ subject to } A\x^0 = F(\x^0)
  \end{align}
where $\mathbf{J}_0$ stands for the jacobian of $f$ at $\x^0$. There is a closed form solution for this problem:
\begin{equation*}
    \mathbf{A} = \mathbf{J}_0 + \frac{1}{\|\x^0\|^2}\left( F(\x^0) - \mathbf{J}_0\x^0 \right)(\x^0)^T.
\end{equation*}

This is, in some sense, the best possible initialization: If $f$ is indeed linear, this initialization implies that our network starts out as the analytical solution to the problem. Moreover, the Hartman-Grobman Theorem indicates that this initialization
becomes increasingly better as $\x^0$ gets close to the established equilibrium point.

However, for nonlinear systems, this naive implementation leads to problems. More specifically, it is possible that $\mathbf{J}_0$ contains eigenvalues with positive real part: in this case, this initialization will lead to a solution that blows up exponentially in time, which is rarely a good first guess.

In these cases, we adapt the approach above. We decompose the matrix $\mathbf{A}$ computed above into its symmetric and  skew-symmetric components. We then keep only the skew-symmetric component, which contains only eigenvalues with zero real part: $\Tilde{\mathbf{A}} = (\mathbf{A} - \mathbf{A}^T)/2$. We then supplement the remaining terms with a constant vector field $\mathbf{b}$ such that:
\begin{equation}
    F(\x^0) = \Tilde{\mathbf{A}}\x^0 + \mathbf{b}.
\end{equation}

\section{Experiments}

Due to space restrictions, we include here the remaining details regarding the experiments in the main paper, as well as additional experiments.

\subsection{Experimental Details - Automatic Differentiation}

We found that using automatic differentiation for the PINNs loss lead to conflict with the numerical solvers underlying the Neural ODEs in Experiment 2. For this reason, all time derivatives were calculated using a second-order, centered finite-difference approximation with small ($10^{-3}$) step size.

Moreover, we found that using the adjoint method for gradient calculations for such small-scale NODEs lead to substantial slowdowns of up to two times. For this reason, gradients were in fact calculated using usual backpropagation, as per the specifications of TorchDyn. In any case, Neural ODEs were still significantly slower than the other two architectures.

\begin{table*}[t]
\footnotesize
\addtolength{\tabcolsep}{-2pt}
  \caption*{Table A. Experiment 3: Lotka-Volterra model}
  \label{tableA}
  \centering
  \begin{tabular}{cccccc}
    \toprule
    Model & &Params   & $\mathcal{L}_{\text{acc}}$ & $\mathcal{L}_{\text{extrap}}$ &  Time(s)\\
    \midrule
    MLP & &$2.3$K & $1.1\tento{0} \pm 1.1\tento{-1}$ & $3.3\tento{1} \pm 7.6\tento{0}$ & $34.1\pm0.6$  \\
    NCF & &$2.5$K & $9.2\tento{-2} \pm 2.3\tento{-2}$ & $3.6\tento{-1} \pm 4.8\tento{-1}$ & $81.8\pm2.7$  \\
    NCF-T & &$2.5$K & $2.6\tento{-2} \pm 4.0\tento{-2}$ & $1.1\tento{-1} \pm 1.8\tento{-1}$ & $82.9\pm1.5   $  \\
    \bottomrule
  \end{tabular}
\end{table*}

\begin{table*}[t]
\footnotesize
\addtolength{\tabcolsep}{-2pt}
  \caption*{Table B. Experiment 4: FitzHugh-Nagumo model}
  \label{tableB}
  \centering
  \begin{tabular}{cccccc}
    \toprule
    Model & &Params   & $\mathcal{L}_{\text{acc}}$ & $\mathcal{L}_{\text{extrap}}$ &  Time(s)\\
    \midrule
    MLP & &$2.3$K & $4.4\tento{-5} \pm 2.1\tento{-5}$ & $5.1\tento{0} \pm 1.8\tento{0}$ & $33.1\pm0.6$  \\
    NCF & &$2.5$K & $1.4\tento{0} \pm 8.8\tento{-3}$ & $1.7\tento{0} \pm 1.9\tento{-2}$ & $80.7\pm1.6$  \\
    NCF-T & &$2.5$K & $1.4\tento{0} \pm 8.8\tento{-3}$ & $2.3\tento{0} \pm 8.9\tento{-2}$ & $80.2\pm1.2   $  \\
    \bottomrule
  \end{tabular}
\end{table*}

\subsection{Experimental Details - FitzHugh-Nagumo}

 The parameters were chosen as $a = b = I = 0$ and $\epsilon=1$ for simplicity, while leading to no significant changes in the system's qualitative behavior. 
 
 For the initial conditions, we chose $\x^0 = (2,-2/3)$ to begin near the limit cycle.

The optimizer was set up with Learning rate $ \alpha = 1\tento{-3}$ and decay parameters $\beta = (0.9,0.99)$. 

\subsection{Experimental Details - Hodgkin-Huxley}

As discussed, the data for the second experiment is synthetic, generated from numerical integration of Hodgkin and Huxley's original model. Their model for the gating variables $n,m,h$ was as follows: 
\begin{equation}
    {\frac {dn}{dt}} = \alpha _{n}(V_{m})(1-n)-\beta _{n}(V_{m})n
\end{equation}
\begin{equation}
    {\frac {dm}{dt}}=\alpha _{m}(V_{m})(1-m)-\beta _{m}(V_{m})m
\end{equation}
\begin{equation}
    {\frac {dh}{dt}}=\alpha _{h}(V_{m})(1-h)-\beta _{h}(V_{m})h,
\end{equation}

Likewise, the nonlinear functions $\alpha$ and $\beta$ are the same as the ones used in the original paper:
$$
\alpha _{n}(V_{m})={\frac {0.01(10-V)}{\exp {\big (}{\frac {10-V}{10}}{\big )}-1}}
$$
$$\alpha _{m}(V_{m})={\frac {0.1(25-V)}{\exp {\big (}{\frac {25-V}{10}}{\big )}-1}}
$$
$$
\alpha _{h}(V_{m})=0.07\exp {\bigg (}-{\frac {V}{20}}{\bigg )}
$$
$$
\beta _{n}(V_{m})=0.125\exp {\bigg (}-{\frac {V}{80}}{\bigg )}
$$
$$\beta _{m}(V_{m})=4\exp {\bigg (}-{\frac {V}{18}}{\bigg )}
$$
$$
\beta _{h}(V_{m})={\frac {1}{\exp {\big (}{\frac {30-V}{10}}{\big )}+1}}
$$

For the real system, the magnitude of these variables varies widely, leading to some currents being much larger than others. When training, this leads to problems, as these differences in magnitude lead to a bias towards minimizing the losses associated to large currents. To minimize this, and to avoid saturating input neurons, we rescale $V,n,m,h$ by factors of $0.1,10,10,10$ respectively. We again initialize the system close to the limit cycle.

The optimizer was set up with learning rate $\alpha = 2.5\tento{-3}$ and decay parameters $\beta = (0.9,0.95)$.


\subsection{Additional Experiment: Causality}

To illustrate the difficulties PINNs have with causality and spurious convergence to trivial solutions, we include an additional experiment. The Lotka-Volterra system, originally built to model ecological competition, has also found applications as an alternative model for spiking neurons (Noonburg 1989):
\begin{equation}
    \begin{aligned}
 {\frac {dx}{dt}}&=\alpha x-\beta xy,\\{\frac {dy}{dt}}&=-\gamma y+\delta xy,
 \end{aligned}
\end{equation}
 
This oscillator, albeit very simple, can be remarkably difficult for PINNs to solve, due to the fact that its orbits veer very close to the equilibrium lines $x=0$ and $y=0$. As a result, regular PINNs often find it easier to gravitate towards this equilibrium.

For this experiment, we measure how often each architecture converges to an incorrect solution, using the same setup as experiment 1. We take $\alpha = \beta = \gamma = \delta = 1$ for simplicity.

The results, given in Table A, clearly indicate that regular PINNs were entirely unable to approximate the solution to this problem (see Figure 1), although they performed very well in Experiment 1. 

Meanwhile, NCFs converged to the correct, physically meaningful solutions. We believe that this is due to the fact that group property II of flows implies uniqueness.

\subsection{Additional Experiment: Nonlinearity}

To illustrate the difficulties of Affine Neural Conjugates with strongly nonlinear phenomena, we run a second instance of Experiment 1, now with an initial condition much closer to the equilibrium at the origin. The orbit starts as an oscillator increasing in amplitude, until it converges to a limit cycle (see Figure A).

\begin{figure}[h!]
  \centering
  \includegraphics[width=.75\linewidth]{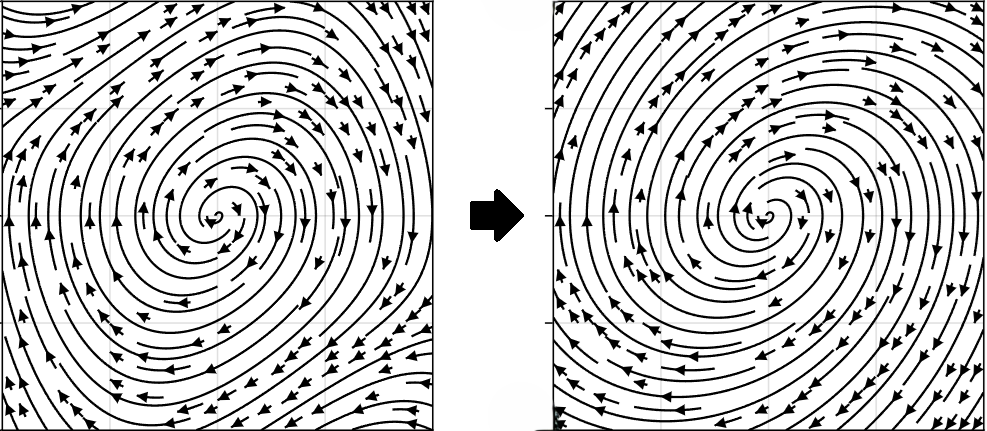}
  \caption*{Figure A: The FitzHugh-Nagumo model and its linearization are locally conjugated near the origin.}
  \label{fig:Banhoeffer}
\end{figure}

This poses a significant challenge to affine NCFs because the system behaves qualitatively similar to two distinct linear systems at different points in time:
At $t=0$, it should behave as a hyperbolic linear system, spiralling outwards (by the Hartman-Grobman Theorem). Later, when it converges to the limit cycle, the system behaves as an harmonic oscillator, with purely imaginary eigenvalues. 

Because the flow $\Psi$ is fixed in time, it cannot adapt to these different realities. As a result, the network can't choose between the two distinct topologies and fails to converge.

\begin{figure}[h!]
  \centering
  \includegraphics[width=.95\linewidth]{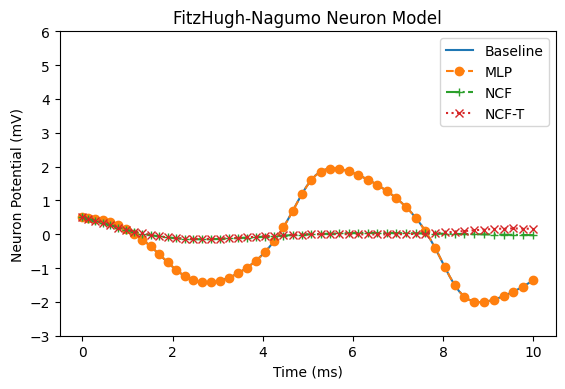}
  \caption*{Figure B. Affine Neural Conjugates may not be able to deal with convergence towards a limit cycle.}
  \label{fig:Nonlinear_fail}
\end{figure}

We emphasize that this is a limitation imposed by the simple topology of affine flows. A more representative flow $\Psi$, although almost certainly more computationally expensive, could mitigate these issues.

On the other hand, the implicit topological regularization imposed by affine flows may, in some contexts, be desirable. In the context of Continuous Normalizing Flows, in which Neural ODEs are often used, a simplified topology should be often enough, and even preferable. Indeed, one is most likely to avoid the appearance of strange attractors and limit cycles in these contexts.

\end{document}